\documentclass{article}
\usepackage{graphicx} 
\usepackage{authblk}

\usepackage[utf8]{inputenc}
\usepackage{graphicx}
\usepackage{fancyhdr}
\usepackage{float}
\usepackage[margin=2cm]{geometry}
\usepackage[pdftex,pagebackref=false]{hyperref} 

\usepackage{cite} 
\usepackage{graphicx}
\usepackage{caption}
\usepackage{subcaption}
\usepackage{amsmath}
\usepackage[section]{placeins}
  
\hypersetup{
    plainpages=false,       
    unicode=false,          
    pdftoolbar=true,        
    pdfmenubar=true,        
    pdffitwindow=false,     
    pdfstartview={FitH},    
    pdftitle={PhD Problem},    
    pdfauthor={Will},    
    pdfnewwindow=true,      
    colorlinks=true,        
    linkcolor=blue,         
    citecolor=green,        
    filecolor=magenta,      
    urlcolor=cyan           
}

\title{Learning Velocity-based Humanoid Locomotion: Massively Parallel Learning with Brax and MJX\footnote{This work has been accepted at the CLAWAR 2024 conference in Kaiserslautern, Germany}} 

\author[1]{William Thibault}
\author[1]{William Melek}
\author[1,2]{Katja Mombaur}
\affil[1]{University of Waterloo, Waterloo, ON N2L 3G1, Canada}
\affil[2]{Karlsruhe Institute of Technology, 76131 Karlsruhe, Germany}

\date{June 30, 2024}

\begin{document}
\maketitle
\begin{abstract}
Humanoid locomotion is a key skill to bring humanoids out of the lab and into the real-world. Many motion generation methods for locomotion have been proposed including reinforcement learning (RL). RL locomotion policies offer great versatility and generalizability along with the ability to experience new knowledge to improve over time. This work presents a velocity-based RL locomotion policy for the REEM-C robot. The policy uses a periodic reward formulation and is implemented in Brax/MJX for fast training. Simulation results for the policy are demonstrated with future experimental results in progress.
\end{abstract}
\section{Introduction}
Recent interest in humanoid robots as general purpose robots has lead to a significant increase in humanoid robotics research and development in both industry and academia. A main reason for the interest is because humanoid robots have the ability to perform repetitive and dull tasks in human environments. A core skill necessary for many tasks, like moving boxes around a warehouse, is robust locomotion. Locomotion planning and control algorithms vary greatly from linear inverted pendulum walking (LIPM) \cite{kajita2003} to online whole-body MPC walking \cite{dantec2022}. Reinforcement learning (RL) has also been a method of choice recently for robotic motion generation given its ability to adapt to different environments or conditions and generalize well to many scenarios. Additionally, RL is a data-driven approach allowing it to improve as it is exposed to new experience. New RL frameworks that optimize GPU usage have enabled faster train times allowing for robust policy learning with domain randomization in mere hours \cite{caluwaerts2023,rudin2022}.
A method for generating a controlled, periodic bipedal gait on Cassie was developed by Siekmann et al. \cite{siekmann2021} through the use of a periodic reward composition where the rewards were provided using an indicator function for the robot matching the gait phase, otherwise it provides penalties. Singh et al. \cite{singh2022} proposed a method for learning planned 3D footsteps in simulation for larger, heavier robots like HRP-5P leveraging a similar periodic reward composition. This work was followed by a current controlled approach for using the trained policy on HRP-5P for planar walking with domain randomization to bridge the sim to real gap \cite{singh2023}.
In this work, a humanoid locomotion RL policy for 2D velocity-based locomotion is proposed for the REEM-C robot. This work uses the periodic reward composition previously mentioned while leveraging faster train times than the previous work through the use of the recent Brax and MJX RL training framework \cite{brax_mjx}. Simulation results for the walking policy are presented in this work along with plans for experimental testing on the real REEM-C robot.
\section{Reinforcement Learning Problem}
\subsection{REEM-C Humanoid Robot}
The RL problem developed in this work is focused on joystick-style velocity locomotion for the full-size humanoid robot REEM-C. This humanoid robot has 30 degrees of freedom with 6 degree of freedom legs, 7 degree of freedom arms, 2 degrees of freedom for the torso and 2 degrees of freedom for the head. In the wrists and feet of the robot are 6D force-torque sensors. To simplify the learning problem, the upper body joints are held fixed and only the position controlled leg joints, with complete dynamics, are controllable in the MJX simulator.
\subsection{Markov Decision Process}
\begin{figure*}[h]
    \centering
    \includegraphics[width=0.8\linewidth, trim={0cm 0cm 0cm 0cm},clip]{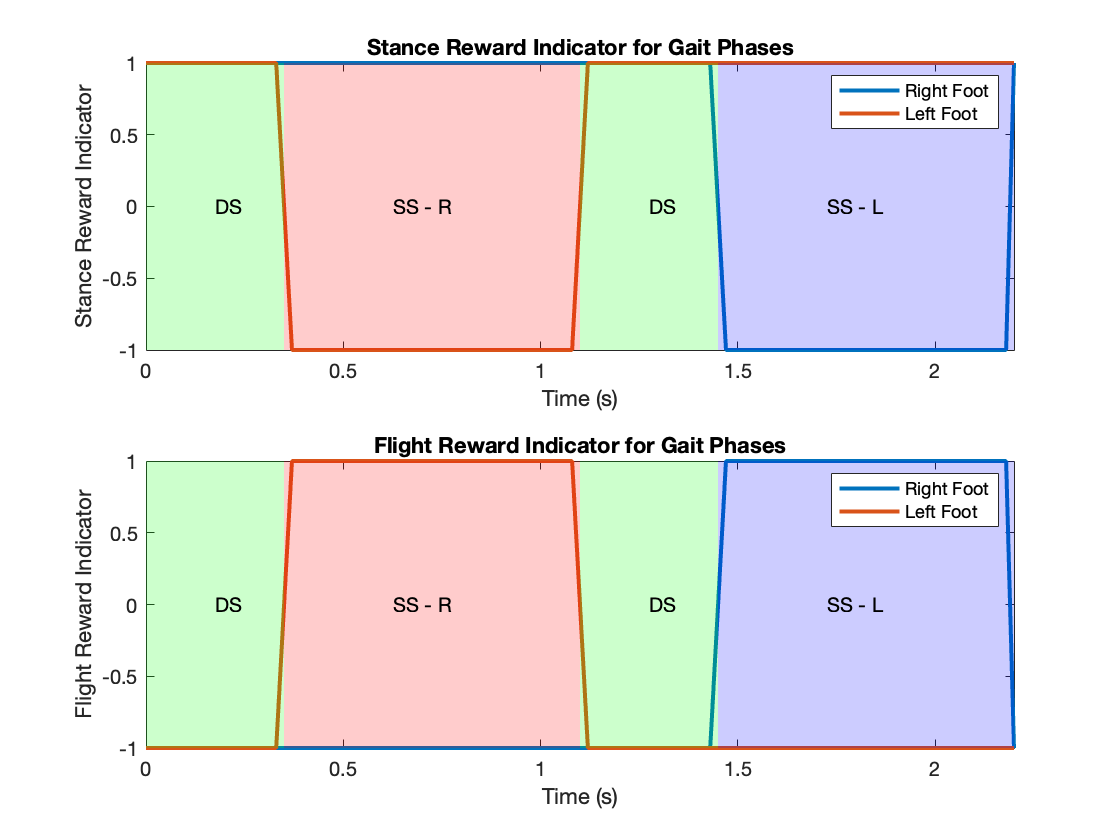}
   \caption{Stance and flight reward indicators depending on different gait phases}
    \label{fig:gait_phases}
\end{figure*}

The Markov Decision Process for this work has a 12 dimensional action space composed of the 12 leg joint positions. The observation space includes the joint state of robot, the periodic clock signal as defined in Singh et al. \cite{singh2022} for a gait cycle with a double support time of 0.35 s and a single support time of 0.75 s, a command velocity of $(x,y,\psi)$. The input velocity has a range of $(x,y,\psi) = ([-0.3,1.0]m/s,[-0.3,0.3]m/s,[-0.5,0.5]rad/s)$, based on reasonable limits and existing joystick LIPM walking for REEM-C. Table \ref{tab:obs_space} summarizes the observation space.
\begin{table}
\caption{Summary of observation space}
\centering
\begin{tabular}{ |p{4cm}|p{4cm}|p{4cm}|}
 \hline
 State & Dimensions & Scaling \\
 \hline
 Yaw Rate & 1 & 0.25 \\
 \hline
 Projected Gravity of Base & 3 & -\\
 \hline
 Command & 3 & [2.0,2.0,0.25] \\
 \hline
 Joint Angles & 12 & - \\
 \hline
 Joint Velocities & 12 & - \\
 \hline
 Last Action & 12 & - \\
 \hline
 Periodic Clock Signal & 2 & - \\
 \hline
\end{tabular}
\label{tab:obs_space}
\end{table}
The rewards include the following rewards or penalties:
\begin{enumerate}
    \item Center of mass linear velocity tracking reward:\\
    $r_1 = 3 * e^{-((\dot{x}_{input}-\dot{x}_{base})^2 + (\dot{y}_{input}-\dot{y}_{base})^2/\sigma}$, where $\sigma$ is a scaling factor
    \item Center of mass angular velocity tracking reward:\\
    $r_2 = 3 * e^{-((\dot{\psi}_{input}-\dot{\psi}_{base})^2/\sigma}$, where $\sigma$ is a scaling factor
    \item Linear z velocity penalty:\\
    $r_3 = -2 * (\dot{z}_{base})^2$
    \item Torso roll and pitch rate penalty:\\
    $r_4 = -0.1 * (\dot{\phi})^2 + (\dot{\theta})^2$
    \item Non-zero torso roll and pitch penalty:\\
    $r_5 = -10 * (\phi)^2 + (\theta)^2$
    \item Torque penalization:\\
    $r_6 = -0.0002 * \sqrt{\Sigma \tau^2} + \Sigma |\tau|$
    \item Action rate penalization:\\
    $r_7 = -0.01 * \Sigma(A_{t}-A_{t-1})^2$
    \item Stand still no motion penalty: \\
    $r_8 = -0.5 * \Sigma|q-q_{default}|$, where this penalty is only included if the normalized velocity command is less than 0.1.
    \item Early termination penalty:\\
    $r_9 = -1$, when the base height drops below 0.7 m, joint angles are exceeded or the robot is falling.
    \item Gait foot contact stance reward or penalty: \\
    $r_{10} = 0.02 * (I_{stance}(F_{z,right}) + I_{stance}(F_{z,left}))$, where the indicator function $I_{stance}$ is defined by the gait phase as shown in Figure \ref{fig:gait_phases}.
    \item Gait foot flight velocity reward or penalty: \\
    $r_{11} = 0.2 * I_{flight}(\Sigma (v_{right})^2) + I_{flight}(\Sigma (v_{left})^2)$, where the indicator function $I_{flight}$ is defined by the gait phase as shown in Figure \ref{fig:gait_phases}.
    \item High foot contact penalty: \\
    $r_{12} = -0.01 * (F_{z,right} + F_{z,left})$, for contact forces exceeding the maximum contact force of 1500 N.
    \item Gait phase reward: \\
    $r_{13} = 1$, when the gait phase changes to promote taking steps.
    \item Velocity rate penalty: \\
    $r_{14} = -0.001 * \Sigma (\dot{q}_n - \dot{q}_{n-1})^2 + \Sigma (\dot{q}_n - 2.0 * \dot{q}_{n-1} + \dot{q}_{n-2})^2$
    \item Pose bias penalty: \\
    $r_{15} = -0.4 * \Sigma (q - q_{default})^2$, where $q_{default}$ is the starting pose.
    \item Head to base projection penalty: \\
    $r_{16} = -2 * ((x_{base}-x_{head})^2 + (y_{base}-y_{head})^2)$
    \item Base height penalty: \\
    $r_{17} = -100 * (z_{base}-0.8)^2$, where 0.8 m is the reference base height.
\end{enumerate}
\noindent The reward functions and weights were largely inspired by Singh et al. \cite{singh2022}, Radosavovic et al. \cite{radosavovic2023}  and the open-sourced Brax/MJX implementation of the work by Caluwaerts et al \cite{caluwaerts2023}.
\section{Results}
\begin{figure*}[h]
    \centering
    {\includegraphics[width=0.19\linewidth, trim={2cm 0cm 2cm 0cm}, clip]{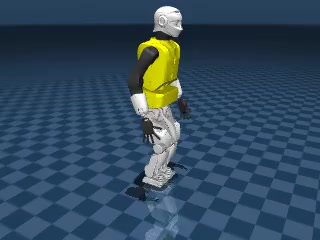}}
    \hfill
    {\includegraphics[width=0.19\linewidth, trim={2cm 0cm 2cm 0cm}, clip]{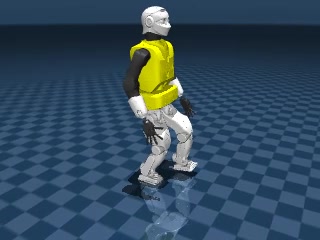}}
    \hfill
    {\includegraphics[width=0.19\linewidth, trim={2cm 0cm 2cm 0cm}, clip]{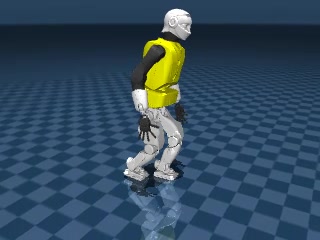}}
    \hfill
    {\includegraphics[width=0.19\linewidth, trim={2cm 0cm 2cm 0cm}, clip]{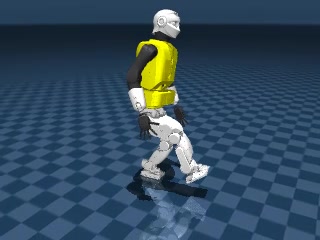}}
    \hfill
    {\includegraphics[width=0.19\linewidth, trim={2cm 0cm 2cm 0cm}, clip]{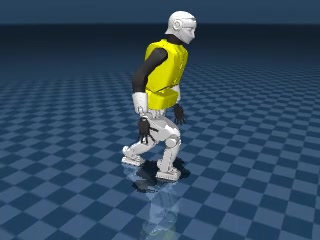}}
    \caption{REEM-C walking forward at 1.0 m/s}
    \label{fig:sim}
\end{figure*}

The RL algorithms of Brax using the MJX simulator, a version of MuJoCo that supports the XLA compiler for GPU and TPU based simulation using JAX, is used to train the locomotion policy. This training pipeline was selected due to the highly parallel approach that can exploit the GPU for fast physics simulation steps and in turn shorter training times. The training is run for 200,000,000 training steps with 8192 parallel environments using PPO. The networks for the PPO algorithm had 4 hidden layers of 128 neurons. This training was performed on a PC with an AMD 7950x CPU, 64 GB RAM, and an RTX 4090 24 GB VRAM and completed training in approximately 56 minutes. Note that to achieve fast and high performance training an optimized model of the robot was used, containing only lower body collisions and recommended MJX performance tuning techniques \cite{mujoco_mjx}. A sample motion sequence of the walking policy can be see in Figure \ref{fig:sim}.
With these promising simulation results, plans for real hardware experiments are in progress. Mainly, the sim to real gap is being closed through better actuator modelling for the simulation and a task space inverse dynamics approach \cite{tsid} for tracking the joint angles, center of mass position and foot contacts.
\section{Discussion}
This work presents initial results of a velocity-based RL locomotion policy for the REEM-C humanoid robot. The RL problem formulation is based on a periodic reward formulation to enforce a periodic stepping gait and is implemented in Brax/MJX for fast training performance, the first implementation for a real bipedal robot to the best knowledge of the authors. Preliminary results produced a stable, periodic walking policy at different command velocities and future hardware experiments will demonstrate the robustness and versatility.
%
%
\bibliographystyle{splncs03}
\bibliography{references.bib}
\end{document}